%% file: main.tex
\documentclass{article}

\usepackage{microtype}
\usepackage{graphicx}
\usepackage{subcaption}

\usepackage{booktabs} 

\usepackage{ amsmath }
\usepackage{amssymb}
\usepackage{ulem}
\usepackage[bottom]{footmisc}

\usepackage{hyperref}

\usepackage[accepted]{icml2019}

\icmltitlerunning{Learning Novel Policies For Tasks}

\begin{document}

\twocolumn[
\icmltitle{Learning Novel Policies For Tasks}



\icmlsetsymbol{equal}{*}

\begin{icmlauthorlist}
\icmlauthor{Yunbo Zhang}{a}
\icmlauthor{Wenhao Yu}{a}
\icmlauthor{Greg Turk}{a}

\end{icmlauthorlist}

\icmlaffiliation{a}{School of Interactive Computing, Georgia Institute of Technology, USA}

\icmlcorrespondingauthor{Yunbo Zhang}{yzhang3027@gatech.edu}
\icmlcorrespondingauthor{Wenhao Yu}{wenhaoyu@gatech.edu}
\icmlcorrespondingauthor{Greg Turk}{turk@cc.gatech.edu}

\icmlkeywords{Machine Learning, ICML, Reinforcement Learning}

\vskip 0.3in
]



\printAffiliationsAndNotice{} 

\input{defs.tex}
\input{Abstract.tex}
\input{Introduction.tex}

\input{Related_Work.tex}
\input{Preliminaries.tex}

\input{Methods.tex}

\input{Experiments.tex}

\input{Discussion.tex}
\input{Conclusion.tex}
\input{Acknowledgement.tex}
\clearpage
\bibliography{bibliography}
\bibliographystyle{icml2019}
\clearpage

\twocolumn[
\icmltitle{Supplementary Materials}









]

\input{zAppendix.tex}

\end{document}

%% file: defs.tex

\newcommand{\cmt}[1]{}
\newcommand{\greg}[1]{\textcolor{blue}{{#1}}}
\newcommand{\wenhao}[1]{\textcolor{cyan}{{#1}}}
\newcommand{\yunbo}[1]{\textcolor{cyan}{{#1}}}
\newcommand{\newtext}[1]{#1}
\newcommand{\original}[1]{\textcolor{magenta}{Original: #1}}
\newcommand{\eqnref}[1]{equation~(\ref{eq:#1})}

%% file: Abstract.tex
\begin{abstract}

In this work, we present a reinforcement learning algorithm that can find a variety of policies (\textit{novel policies}) for a task that is given by a task reward function. Our method does this by creating a second reward function that recognizes previously seen state sequences and rewards those by novelty, which is measured using autoencoders that have been trained on state sequences from previously discovered policies. We present a two-objective update technique for policy gradient algorithms in which each update of the policy is a compromise between improving the task reward and improving the novelty reward. Using this method, we end up with a collection of policies that solves a given task as well as carrying out action sequences that are distinct from one another. We demonstrate this method on maze navigation tasks, a reaching task for a simulated robot arm, and a locomotion task for a hopper. We also demonstrate the effectiveness of our approach on deceptive tasks in which policy gradient methods often get stuck.

\end{abstract}

%% file: Introduction.tex
\section{Introduction}

Deep Reinforcement Learning (DRL) has shown great potential in solving problems in various domains such as game playing, maze navigation, and robotic control.  Often times it is sufficient to find a single policy that solves the given task.  In some cases, however, it may be desirable to find several different policies that solve the problem in different ways.  If, for instance the goal is to produce a locomotion policy for a legged robot, there may be several ways in which the robot may coordinate its limbs, leading to various styles of walking.  Similarly, there might be several ways in which a robot arm can reach a given target, and some of these reaching motions may prove more useful than others in different situations.

The goal of our work is to provide reinforcement learning framework for finding not just a single policy for a given task, but instead find a variety of distinct policies, each of which also solves the task at hand.  We will refer to each of these various policies for a given task as a \textit{novel policy}.  Our method of finding novel policies draws upon the recent success of policy gradient approaches to solving continuous control problems, such as PPO \cite{ppo}, TRPO \cite{trpo}, and DDPG \cite{ddpg}.  A given policy gradient algorithm is guided towards producing a policy by a provided reward function, and the algorithm seeks a policy that maximizes the cumulative reward.  If a particular policy that has been produced by a policy gradient algorithm is somehow insufficient for a given task, or if further variation is desired, there are typically two alternatives.  One possibility is to run the algorithm again using a different random number seed, and hope that this will result in a different policy.  The second option is to modify or augment the reward function so that the policy gradient algorithm is guided towards a different policy.  Our method offers a different approach to finding novel policies that does not rely on chance or reward tuning.

We formulate the problem of finding novel policies as a multi-objective optimization problem.  First, we create an autoencoder that recognizes the sequence of states that previous policies typically generate.  We use this autoencoder to create a function that rewards novelty, that is, that rewards sequences of actions that are significantly different than those of the previous policies.  We then solve a reinforcement learning problem in which there are two objectives: 1) solve the given task (guided by the task reward function), and 2) perform sequences of actions that are novel when compared to previous policies (guided by a novelty reward function).  There are several ways in which a policy gradient algorithm may solve such a multi-objective problem.  One way is to create a new reward term that is simply a weighted sum of the task reward and the novelty reward.  Another possibility is to examine the policy gradient with respect to each of these reward functions, and update the policy in a way that attempts to follow both these gradients.  We explore both these methods, and in particular we demonstrate that using an angle bisector between the task and novelty gradients is an effective approach across a variety of tasks.

We demonstrate our approach on several problems, including a maze with multiple goals, a reacher task for a simulated robot arm, and a hopper.  We also demonstrate the application of our approach to two deceptive problems (deceptive maze, deceptive reacher), for which most algorithms get stuck in a poor local minimum.  Our approach finds policies that avoid the behaviour of earlier failed policies, thus making it easier to learn a policy that achieves the given task.

In this work we make the following contributions:

\begin{itemize}

\item{We formulate the problem of \textit{finding novel policies}, in which the goal is to sequentially produce policies that both solve a given task and that do so in a manner that is distinct from prior policies.}
\item{We show how to measure the novelty of a given policy's behavior using the output of an autoencoder that examines the recent history of a policy.  
This measurement of novelty can then be used to find multiple novel policies for a given task.}
\item{We present a method called the Task-Novelty Bisector (TNB) that produces novel policies that solve a given task.  We demonstrate TNB when it is used together with PPO, but our method is applicable to any policy gradient method.}

\end{itemize}

%% file: Related_Work.tex
\section{Related Work}

Our method builds on top of the basic policy gradient approach for reinforcement learning \cite{sutton2018reinforcement}. There have been  a number of recent policy gradient algorithms that have been used to create neural network policies for continuous control such as PPO \cite{ppo}, TRPO \cite{trpo}, DDPG \cite{ddpg}.

Our work has some relation to the problem of finding \textit{options} \cite{optionFramework}.  An \textit{option} is a course of actions that are determined by a policy.  Usually options are used in a hierarchical setting, in which a high level policy selects from a variety of possible options over the course of solving a higher level problem. Some prior works have studied on option learning \cite{vic,option_critic,hausman2018learning}. In particular,  \citet{vic} maximizes the mutual information between options and the final states of a trajectory to encourage the discovery of different intrinsic options. \citet{option_critic} uses a hierarchical structure of policies in which options are created by learning a set of intra-option policies, termination functions, and an policy over options simultaneously. However, neither of these methods have been used to solve continuous robotic control problems.  



Some other work on options, including DIAYN \cite{diayn} and VALOR \cite{valor}, address problems in robotics control. DIAYN \cite{diayn} encourages the diversity of a policy by maximizing the mutual information between options and states while minimizing the mutual information between options and actions conditioned on states. VALOR \cite{valor} uses an LSTM variational autoencoder for option discovery where a universal policy conditioned on an option serves as the encoder, and a bidirectional LSTM serves as the decoder. Neither DIAYN nor VALOR use a task-relevant reward term, and they solely maximize behaviour discovery. Both methods aim to create a set of policies that can be used as initial policies for later task relevant training, often to be used as a low-level controller in a hierarchical setting. In the same spirit of encouraging diversity over an entire trajectory, work from the evolutionary algorithms community has applied Novelty Search \cite{evoNovel,lehman2011evolving,pugh2016quality,mouret2009overcoming} to solve various problems.

Our work is also related to encouraging exploration in reinforcement learning. Some research uses the approach of curiosity-driven exploration \cite{schmidhuber1991curious,sun2011planning,conti2018improving} and applies these techniques to reinforcement learning \cite{Vime,hester2017intrinsically}. For example, to encourage exploration, VIME \cite{Vime} uses variational inferences to derives an intrinsic reward objective that maximizes the information gain about the environment dynamics. \citet{sac,haarnoja2017reinforcement,haarnoja2018soft} add an entropy term in the objective function so that exploration is encouraged through maximizing entropy for visited states. GEP-PG \cite{colas2018gep} tried to decouple exploration and exploitation by focus in one or the other in stages to encourage exploration. Although these methods encourage policy discovery with more exploration, they are not used to intentionally generate multiply distinct policies that solve a given task.



%% file: Preliminaries.tex
\section{Preliminaries}
\subsection{Deep Reinforcement Learning}
Deep Reinforcement Learning solves problems that are modeled as Markov Decision Processes (MDPs), defined by a tuple $\left( \mathcal{S},\mathcal{A},\mathcal{T},r,\rho_{0},\gamma \right)$, where $\mathcal{S}$ is the state space, $\mathcal{A}$ is the action space, $\mathcal{T}:\mathcal{S} \times \mathcal{A} \mapsto S$ is the transition function, $r:\mathcal{S} \times \mathcal{A} \mapsto \mathbb{R}$ is the reward function, $\rho_{0}$ is the initial state distribution, and $\gamma \in \left[ 0,1\right]$ is a discount factor. The goal of RL is to find a policy $\pi_{\theta} : \mathcal{S} \mapsto \mathcal{A}$ parameterized by $\theta$ that maximizes the expected return:
\begin{equation}
\label{eq:rl_obj}
J(\theta) = \mathbb{E}_{s_0, a_0, \dots, s_T}\left[\sum_{t=0}^{T}\gamma^{t}r(s_t, a_t)\right],
\end{equation}
where $s_0 \sim \rho_0(\cdot)$, $a_t = \pi_\theta(s_t)$ and $s_{t+1} = \mathcal{T}(s_{t}, a_{t})$.

\subsection{Policy Gradient Methods}
Policy gradient methods are a class of RL algorithms that estimate the gradient of the expected return with respect to the policy parameters $\theta$. The gradient estimation of the return $R(\theta)$ can be expressed as: 
\begin{equation}
\label{eq:pg}
\frac{\partial J}{\partial \theta} = \mathbb{E}_{s_0, a_0, \dots, s_T}\left[\sum_{t=0}^{T}\nabla\log{\pi_{\theta}(a_t | s_t)}Q(s_t, a_t)\right],
\end{equation}
where $Q(s_t, a_t)$ is the Q function that measures the expected return of taking action $a_t$ at $s_t$ and following the policy $\pi_\theta$ thereafter. 

%% file: Methods.tex
\section{Methods}
\label{sec:methods}

In this work, we propose an algorithm that incrementally build a set of \textit{novel policies} that solve a given task while exhibiting distinct behaviors to each other. At each iteration of our method, we train a new policy for which the learning agent is rewarded for both solving the task and demonstrating novel behaviors compared to the previously trained policies. We define novelty as dissimilarity between state sequences visited by the policies, and this is computed using autoencoders represented by neural networks. To achieve balanced learning progress between solving the task and seeking novel behavior, we propose a two-objective update method for policy gradient algorithms. After the policy is trained, it is expected to solve the task while also being novel in comparison to all of the previous policies. We then train an autoencoder for this newly generated policy to recognize behaviors similar to it and update the autoencoder set. This process is repeated until the desired amount of novel policies have been trained.


\subsection{Measuring Novelty}

Given a rollout generated by the current policy of interest, we want to measure how novel it is compared to rollouts generated from previously trained policies. To do this, we need to define a metric for measuring novelty between rollouts. One possible approach would be using nearest-neighbour based approch as in \cite{zhao2009anomaly,liao2002use} to measure the difference between the newly generated rollout and the rollouts from previous policies and aggregate them to obtain a novelty measurement. However, such methods can be computationally expensive as more rollouts are generated for comparison. Alternatively, one can use a count based novelty detection as as in \cite{bellemare2016unifying,tang2017exploration,strehl2005theoretical} to measure how frequently each state is visited by previous policies and compare that to the new rollout. However, this approach is difficult to scale to problems with high dimensional continuous state space such as the ones in our experiments. 

To generalize across a variety of problems, using neural network for novelty detection as in \cite{hawkins2002outlier,richter2017safe,ruff2018deep} becomes a more desired approach. In this work, we train autoencoders to measure the novelty of a rollout given a set of existing rollouts. The key idea is that if an autoencoder is trained on data from a particular distribution, it will be good at reconstructing data from that distribution, while it will perform poorly if the data is from a different distribution, i.e. when the data is novel as compared to the training data. To measure novelty of a rollout, we train one autoencoder for each trained policy in the sequence. There are two potential pitfalls if we were to use just a single autoencoder for all of the policies that have been found so far. First, if we used a single autoencoder, we would need to retain or regenerate the training rollouts for the earlier policies in the sequence when a new policy is trained, in order that the autoencoder does not forget the rollouts from previous policies. Second, when a single autoencoder is trained on data from multiple policies with distinct behaviors, it may generalize to data that has not been seen in the training data, potentially leading to under-estimated novelty for a novel rollout.

To fully capture the characteristics of the policy, it may be tempting to use the entire rollout as input to the autoencoder. However, such model only measures the novelty at the end of each rollout, leading to a sparse and delayed learning signal. In our method, we instead use a sub-sampled partial rollout as input to the autoencoder. Specifically, during the training of autoencoder, we divide each rollout into multiple fixed-length segments of length $L$ and sub-sample them with a stride of $G$ to obtain the training data, each of length $L/G$. In addition, we use only the states of each segment and ignore the actions, as we are more interested in state sequences that are novel. More details regarding autoencoder training can be found in the supplementary material. 

Given a state sequence $\mathbf{s}=(s_t, s_{t+G}, s_{t+2G}, \dots, s_{t+L})$ and a set of autoencoders $\mathbf{D}=\{\mathcal{D}_i\}$ trained for previous policies, we measure its novelty as:
\begin{equation}
\label{eq:novel_rew}
    r_{novel} = -exp\left(-w_{novel}\min_{\mathcal{D}\in\mathbf{D}}||\mathcal{D}_\pi(\mathbf{s}) - \mathbf{s}||^2\right),
\end{equation}
where the exponential function bounds the range of the novelty reward and $w_{novel} > 0$ modulates the sensitivity of of the reward to the autoencoder reconstruction error (higher $w_{novel}$ means more sensitive). During the training of a new policy, we use Equation \ref{eq:novel_rew} to compute the \textit{novelty reward} for each step in the rollout. Note that for the first $L$ steps we set the novelty reward to zero as there is insufficient data. Once step $L$ has been reached, we can compute the novelty reward, and we have found this to provide a sufficient signal for learning novel behaviors.

\subsection{Two-Objective Optimization}
Since we want to learn policies that solves the task and that behaves differently than previous policies, we want to optimize the two reward functions $r_{task}$ and $r_{novelty}$ simultaneously. A straightforward way to do this would be to use a weighted average of the two reward functions as a single reward. However, this requires fine-tuning of the weights for blending the two reward functions. 

In this work, we propose a two-objective update method to optimize both rewards at the same time, without the need to tune the relative weights. We will refer to our method as the \textit{Task-Novelty Bisector} (TNB) approach.  Using the two reward functions, we can formulate two objective functions $J_{task}(\theta)$ and $J_{novel}(\theta)$:

$$
\begin{aligned}
J_{task}(\theta) =& \mathbb{E}_{s_0, a_0, \dots, s_T}\left[\sum_{t=0}^{T}\gamma^{t}r_{task}(s_t, a_t)\right]\\
J_{novel}(\theta) =& \mathbb{E}_{s_0, a_0, \dots, s_T}\left[\sum_{t=0}^{T}\gamma^{t}r_{novel}(s_t, a_t)\right],
\end{aligned}
$$

and use Equation \ref{eq:pg} to estimate their gradient with respect to the policy parameters:

\begin{equation}
    g_{task} = \frac{\partial J_{task}}{\partial \theta}
\label{eq:g_task}
\end{equation}
\begin{equation}
    g_{novel} =\frac{\partial J_{novel}}{\partial \theta}. \label{eq:g_novel}
\end{equation}

\begin{figure}[h]
\vskip -0.2in
\centering
\begin{subfigure}[h]{0.2\textwidth}
    \centering
    \includegraphics[width=\textwidth]{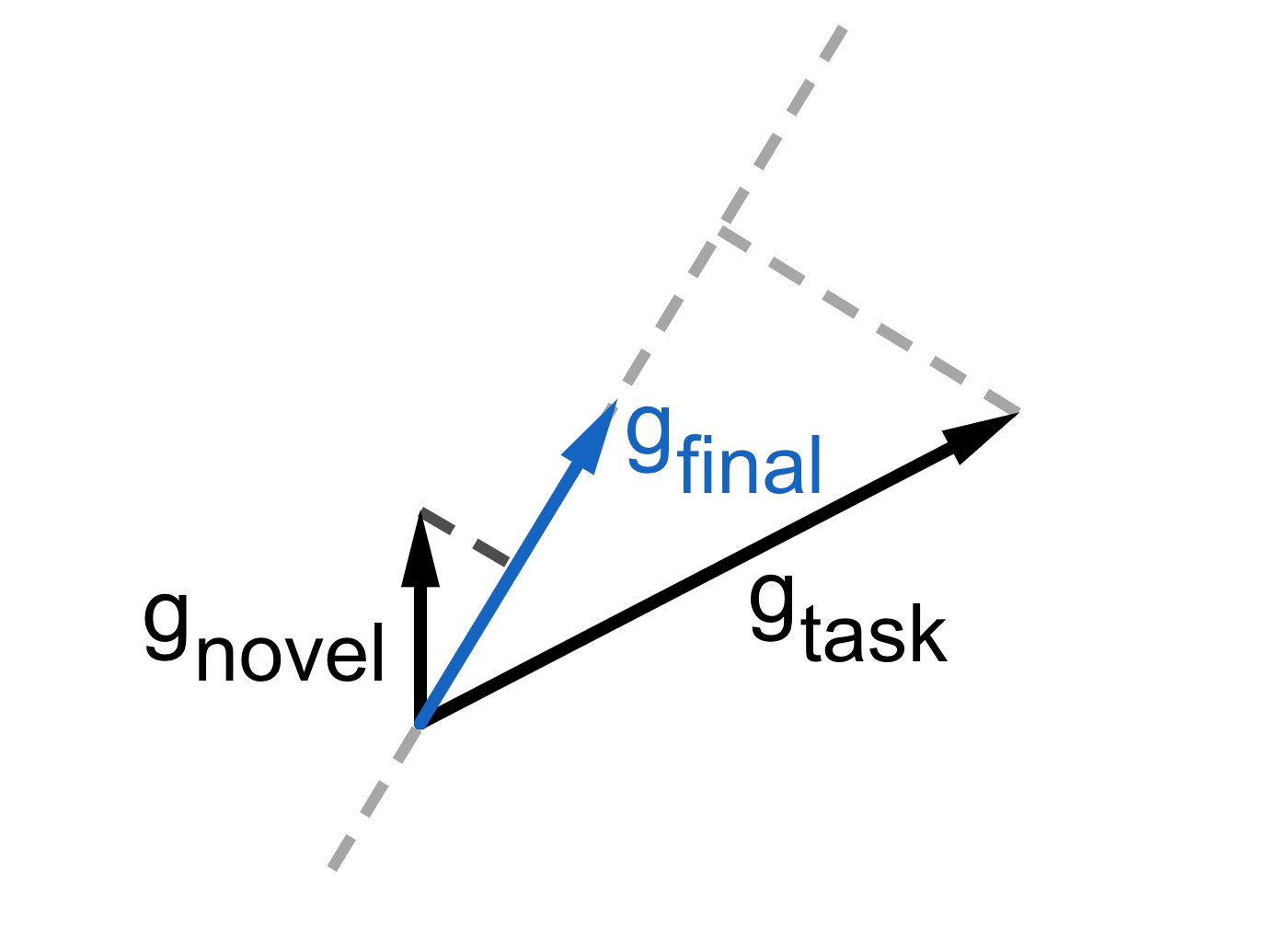}
     \caption{$g_{task}\cdot g_{novel} > 0$}
 \end{subfigure}
\begin{subfigure}[h]{0.2\textwidth}
    \centering
    \includegraphics[width=\textwidth]{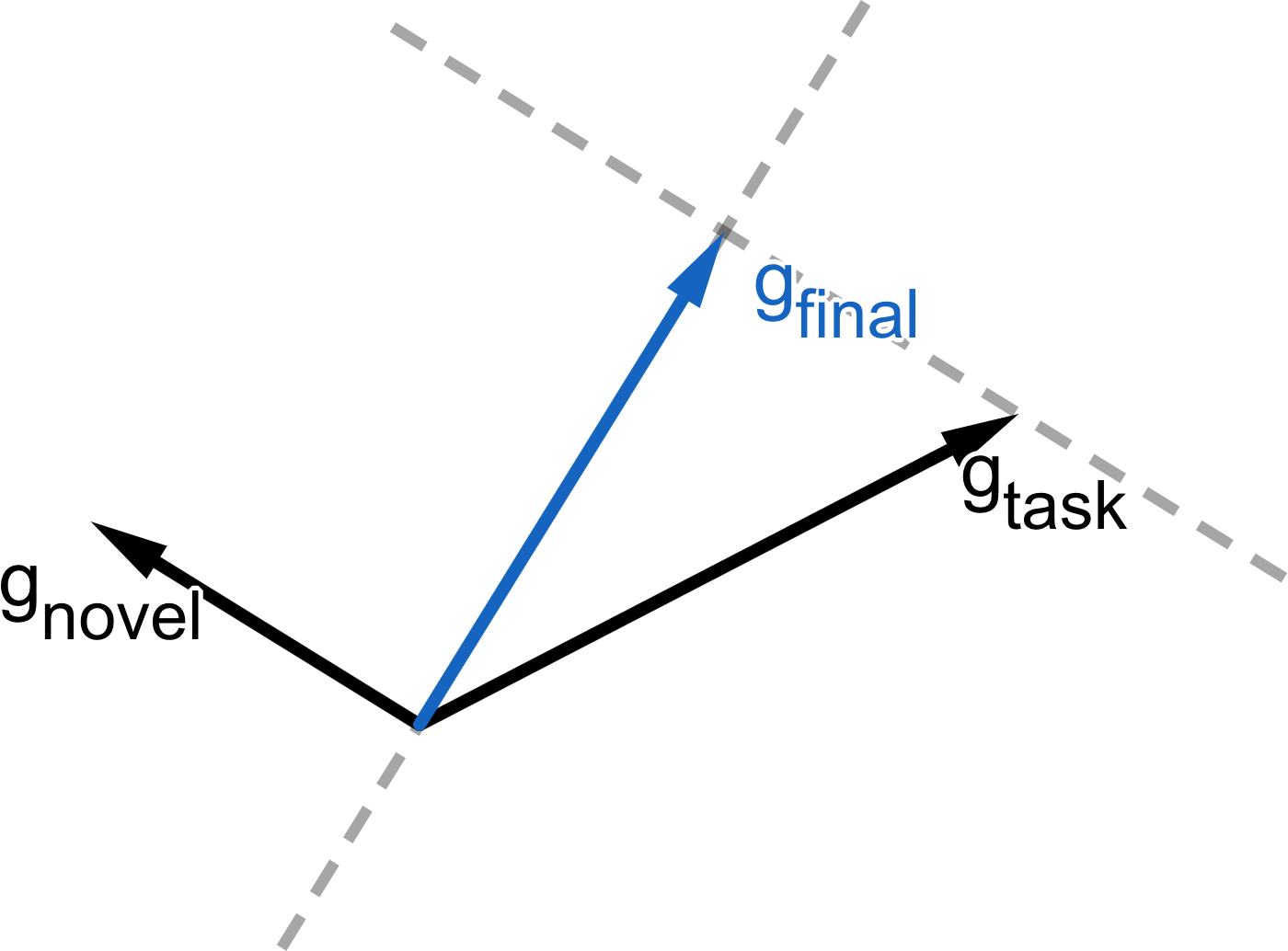}
     \caption{$g_{task}\cdot g_{novel} < 0$}
 \end{subfigure}
 \caption{Final update gradient that improves both the novelty and task objectives.}
\label{fig:gradient_update}
\end{figure}

When taking a weighted average of the two reward functions directly, it is similar to taking the weighted average of the corresponding gradients $g_{task}$ and $g_{novel}$. However, when the two gradients are notably different in magnitude, this may result in updates that are biased toward one reward and thus requires weight tuning. We propose to instead update the policy in the direction of the angular bisector of the two gradients as illustrated in Figure \ref{fig:gradient_update} (a). For calculating the magnitude of the resulting update, we project both gradients onto the direction of the angular bisector and take the mean of the projected gradient magnitude. This results in an update direction that is independent of the scales of the two reward functions and is expected to improve the objective functions of both tasks. This scheme works well when the $g_{task}$ and $g_{novel}$ are pointing in the similar direction. However, when the gradients are pointing in opposite directions, i.e. $g_{task}\cdot g_{novel} < 0$, taking the angular bisector direction may result in an update that improves the objective functions little or not at all for either rewards. To address this issue, we propose a second component to our update scheme. When 
$g_{task}\cdot g_{novel} < 0$, we project $g_{task}$ onto the hyperplane that is perpendicular to $g_{novel}$ and use the projected vector as the final gradient, as shown in Figure \ref{fig:gradient_update} (b). The idea is that when the two gradients do not agree with each other, we want to prioritize solving the task over seeking novel behaviors. Algorithm \ref{alg:TNB_Final_Gradient} shows how we compute the final policy gradient $g_{final}$ from the two gradients and Algorithm \ref{alg:TNPG} describes how we use the proposed method for learning a new policy.



\begin{algorithm}[tb]
   \caption{Task Novelty Policy Learning}
   \label{alg:TNPG}
\begin{algorithmic}[1]
   \STATE {\bfseries Input:} Autoencoders $\textbf{D}=\{\mathcal{D}_1,\mathcal{D}_2,\dotsm,\mathcal{D}_n\}$, Learning rate $\alpha$
   \STATE {\bfseries Initialize:} Policy weights $\theta$
   \FOR{{iteration =} $1,2,\dotsm$}
   \STATE Collect trajectories $\mathbf{\tau}$ using $\pi_\theta$
   \STATE Assign rewards to steps in $\mathbf{\tau}$ using $r_{task}$ and $r_{novel}$
   \FOR{{\bfseries each} epoch}
   \STATE Compute the gradient $g_{final}$ using Algorithm~\ref{alg:TNB_Final_Gradient}
   \STATE $\theta$ = $\theta + \alpha g_{final}$
   \ENDFOR
   \ENDFOR
\end{algorithmic}
\end{algorithm}

\begin{algorithm}[tb]
   \caption{Task-Novelty Bisector Gradient}
   \label{alg:TNB_Final_Gradient}
\begin{algorithmic}[1]
   \STATE {\bfseries Input:} Task policy gradient $g_{task}$ and Novelty policy gradient $g_{novel}$
   \IF {$g_{task}\cdot g_{novel} > 0$}
   \STATE $g_{finalDir} = \frac{bisector(g_{task},g_{novel})}{\|bisector(g_{task},g_{novel})\|}$
     \STATE $g_{final} = \left(\frac{g_{task}+g_{novel}}{2}\cdot g_{finalDir}\right)g_{finalDir}$
   \ELSE
   \STATE $g_{final} = g_{task}-\frac{g_{task} \cdot g_{novel}}{\|g_{novel}\|^2}g_{novel}$
   \ENDIF
\end{algorithmic}
\end{algorithm}



%% file: Experiments.tex
\section{Experiments}
\label{sec:experiments}

\begin{figure*}[h!]
\centering

\includegraphics[width=0.8\textwidth]{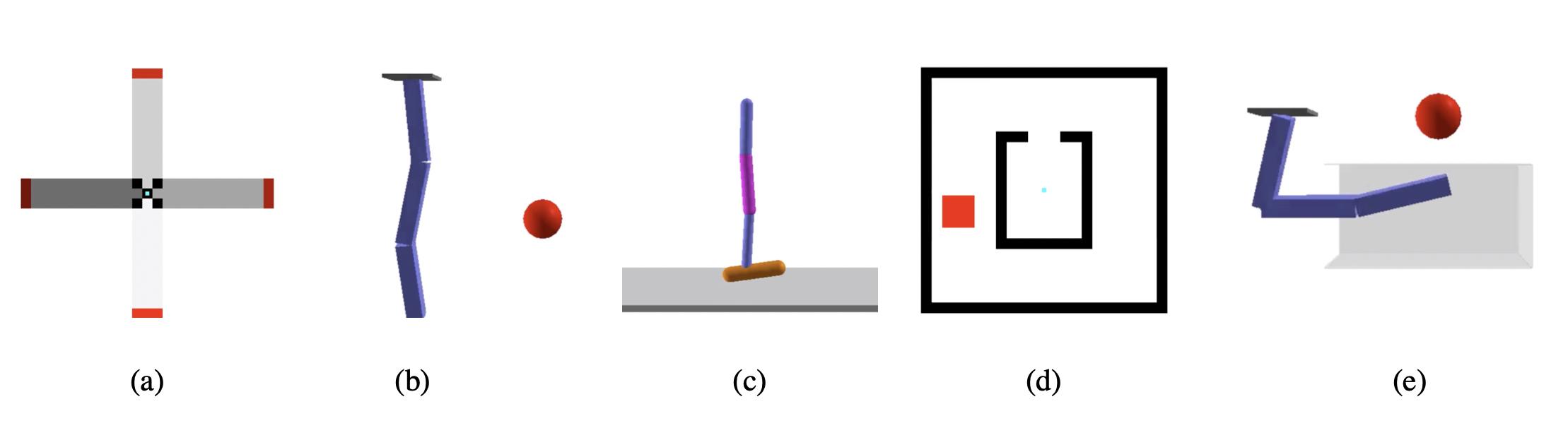}
\vskip -0.2in
\caption{(a) - (e) are 4-Way Maze, Reacher, Hopper, D-Maze, and D-Reacher environments}
  \label{fig:env_init_states}
 \end{figure*}
We use five environments to evaluate our method: 4-Way Maze, Reacher, Hopper, Deceptive Maze (D-Maze), and Deceptive Reacher (D-Reacher). The visualizations of each environment are shown in Figure~\ref{fig:env_init_states}. We provide videos and implementations of these environments at  \href{https://sites.google.com/view/learningnovelpolicy/home}{https://sites.google.com/view/learningnovelpolicy/home}.

\subsection{Implementation Details}

Our TNB method of two-objective update is built on top of the PPO \cite{ppo} implementation in OpenAI Baselines \cite{dhariwal2017openai}. We implement both the 4-Way Maze and D-Maze environments in OpenAI Gym \cite{brockman2016openai}. For the physics simulation of the Reacher, Hopper and Deceptive Reacher, we use the DART physics engine \cite{lee2018dart}. Each rollout for each of the environments has a horizon of 500 control steps unless it triggers an early termination criterion.


For each newly trained policy, we used the final policy and a few policies before convergence to generate the data for the autoencoder. Details are given in the supplementary material. 

\subsection{Algorithm Comparisons}

In the five experiments below, we compare the Task-Novelty Bisector (TNB) algorithm of Section~\ref{sec:methods} to four other methods: 1) regular PPO with different random seeds, 2) Weighted Sum of task and novelty rewards (WSR), and 3) a simplified Task-Novelty Bisector that always uses bisector, and does not perform projections when $g_{task}$ and $g_{novel}$ are pointing in opposite direction (TNB-NoProj). Specifically for the WSR method, we evaluated it on a set of weights and report the best performing one. The detailed results for different weights are given in the supplementary material. We compare to plain PPO with random seeds because, as discussed in  \citet{henderson2017deep}, different random seeds can produce significantly different policies. In addition, for both deceptive problems and the 4-Way Maze problem, we compare our method with Soft-Actor-Critic (SAC) using different random seeds. 

\begin{figure}[b!]
\vskip -0.2in
\centering
\includegraphics[width=0.3\textwidth]{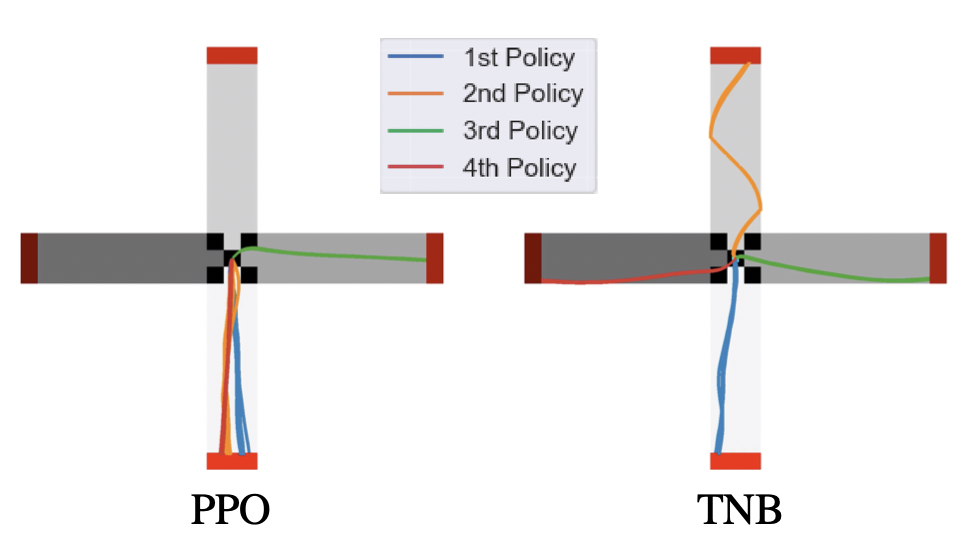}
\vskip -0.1in
\caption{\textbf{4-Way Maze:} Trajectories from a trial of PPO (left) and TNB (right). The colors of the trajectories indicate where in the sequence the corresponding policy is from (first to last): blue, orange, green, red.}
\label{fig:4-Way_Maze_Vis}
\end{figure}

In this section, we use the term \textit{trial} to denote $k$ sequential runs of a given algorithm, i.e. one trial produces $k$ distinct policies. Regardless of method, the first policy is created using regular PPO, without any novelty reward.  For WSR, TNB, and TNB-NoProj, the second policy is trained using a novelty reward based on the first policy. The third policy is trained using a novelty reward from the first two policies, and so on. For these three algorithms, we use the same random seed to initialize all of the $k$ policies for a single trial. In this way, the amount of novelty in a trial cannot be attributed to a difference in random seeds. For a given PPO or SAC trial, we use different random seeds to initialize each of the $k$ policies.

\subsubsection{4-Way Maze}
\begin{table}[t]
\caption{Experiments in 4-Way Maze measure the average number of path explored in each trial, and the number of trials that explore the paths in the order of total rewards.}
\vskip -0.2in
\hskip -0.2in
\begin{center}
\begin{small}
\begin{sc}
\begin{tabular}{lcc}
\toprule
Methods & Avg \# Paths Found & \# Trials In Order \\
\midrule
SAC          & 1& 0/5\\
PPO          & 1.4& 0/5\\
WSR         &3.8 &3/5\\
TNB-NoProj    &\textbf{4} &3/5\\
TNB&\textbf{4} &\textbf{4/5}\\
\bottomrule
\end{tabular}
\end{sc}
\end{small}
\end{center}
\vskip -0.1in
\label{table:4-Way_Maze}
\end{table}
Our first experiment is the 4-Way Maze, a simple 2D navigation environment for a point mass. The agent is placed in the center, and has easy access to the four arms of the maze that each lead to a goal. The observation space of the agent contains the position and velocity of the point mass, and the action space is the 2-dimensional force applied on the point mass.

For calculating the reward, the map is discretized in grid cells, and different color of the cell represents different reward when the agent steps on. Floors are gray-scaled cells and goals are red cells with various brightness. The magnitude of reward signals for a cell depends on the brightness of the color on the cell. Brighter cells will provide more rewards, and darker cells will provide less rewards. The detail of the reward function design are described in the supplementary material. 
The agent receives a one-time small reward when stepping on any floor cells, and it receives a large reward when it steps on a goal cell, followed by the termination of the rollout. 

In the case of the 4-Way Maze, we set $k=4$ so that four policies are created per trial. We run five such trials using different random seeds in order to evaluate each algorithm, and thus create $5 \times 4 = 20$ policies for one algorithm.  We evaluate the diversity of policies using two criteria: 1) The number of paths out of four that are explored in a trial, and 2) Whether the paths in a trial are discovered in descending order of rewards receiving on the path. The first criterion shows whether novel policies are found during training, and the second criterion evaluates whether each subsequent policy is the next best possible solution. 

Figure~\ref{fig:4-Way_Maze_Vis} gives a visual comparison of PPO and TNB, showing several rollouts for each of the four policies that were created using PPO or TNB.  For the trials shown, PPO with random seeds only explores two paths, whereas the TNB policies explore all four paths in order.

Table~\ref{table:4-Way_Maze} shows the results of the 4-Way Maze trials. The values for average number of paths explored shows that both plain PPO and SAC had difficulty finding paths along multiple maze arms. Each of the three variations that use the novelty reward term was able to find all four paths in each trial. The rightmost column shows how many of the five trials found the four possible paths in increasing order of difficulty.  Since PPO and SAC never produced all four paths, they scored zero out of five. In most of the trials, WSR, TNB-NoProj and TNB found all four paths in order, with TNB scoring the highest.



 
\subsubsection{Reacher}

To test our method on a continuous robotic control problem, we use a variant of the classic 3D-Reacher environment.  In this problem, a 3-linked robot arm attempts to move its end effector to a fixed target. The reacher has total of 5-DOFs, one universal joint at the base, one revolute joint connecting the first and second links, and one universal joint connecting the second and the end effector. The observation space is a 26D vector consist of the position of the target, a vector pointing from the arm tip to the target, the sine and cosine of each DOF, and the angle $q$ and angular velocity $\ddot{q}$ for all DOF's. The action space is a 5D vector representing the torque on each of the DOF. The reward function is designed as the distance from the end effector to the target minus the scaled sum of torques on each DOF. In addition, a rollout will be terminated, and a large reward will be added when the reacher touches the target.

\begin{figure}[t!]
\centering
\centerline{\includegraphics[width=0.45\textwidth]{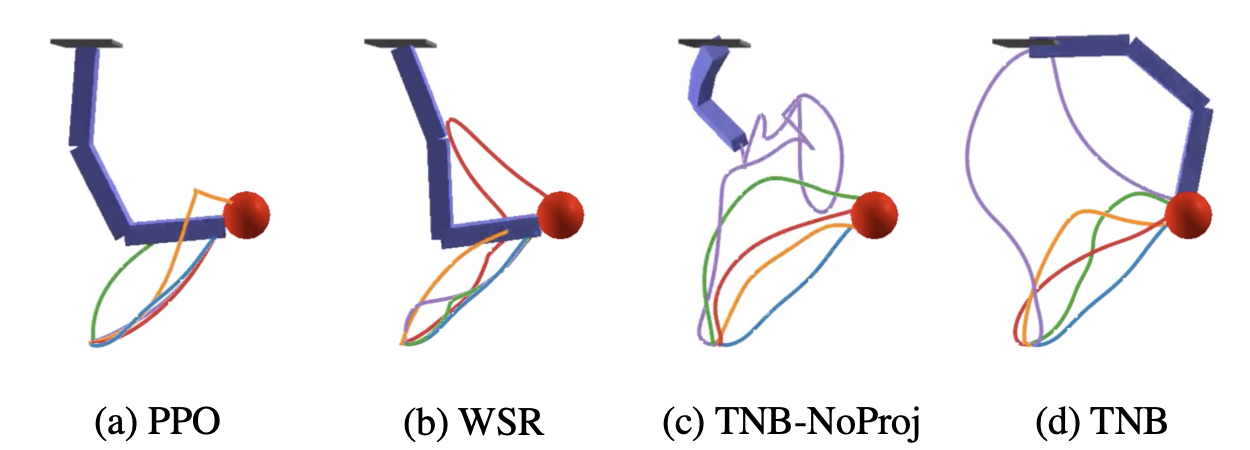}}
\vskip -0.1in
\caption{\textbf{Reacher:} End effector trajectories from a trial for each of the four methods. Policies were trained in this order: blue, orange, green, red, purple.}
\label{fig:Reacher_Vis}
\vskip -0.2in
\end{figure}

For evaluating the reacher environment, we set $k=5$ in each trial for each of PPO, WSR, TNB-NoProj and TNB. As shown in Figure~\ref{fig:Reacher_Vis}, we can judge the novelty of policies visually. To produce this figure, we plot the trail of the reacher's end effector. For different random PPO trials, we see few variations on the trajectories, and in fact three of the policies end up finding nearly the same solution.  WSR and TNB both generate policies that are novel solutions to the reaching task. For TNB-NoProj, on the other hand, some policies fail to solve the task of reaching the target (e.g. the purple curve in Figure~\ref{fig:Reacher_Vis} (c)). This is possibly due to the relative gradient directions between the task reward and the novelty reward being too different. Always using the bisector as gradient update (in TNB-NoProj) may have the effect of too little improvement in task performance, hence leading to a policy that fails to solve the task. See the provided video to view the reacher motions.

\subsubsection{Hopper}

We show that our method is useful for complex locomotion control problems by evaluating our method on the hopper environment. The observation space is a 11D vector containing the linear position and velocity, and all joint angles and velocities. The action space is a 3D vector representing the torque exerted on the three joints. 

\begin{figure}[h]
\centering
\centerline{\includegraphics[width=0.85\columnwidth]{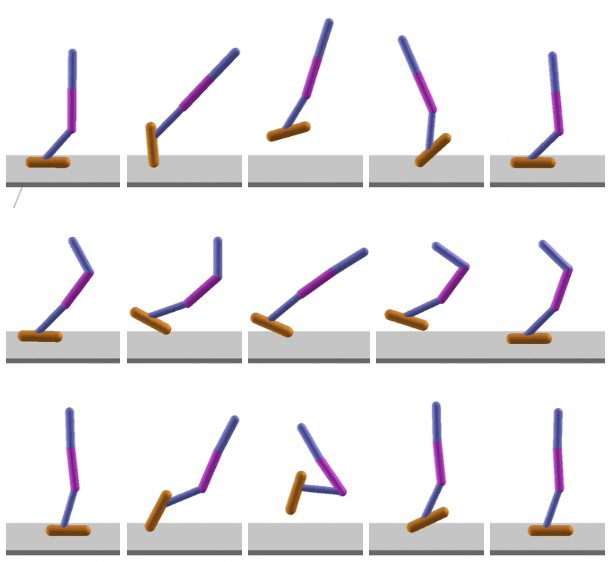}}
\caption{\textbf{Hopper:} The first three policies in a trial sequence when trained using TNB.}
\label{fig:Hopper_Vis}
\end{figure}

We run five trials for each of the methods with $k=5$, which gives us 25 policies. As a result, most policies trained with plain PPO generate trajectories that show very little variation. With novelty reward, on the other hand, the hopper is able to hop forward with variation of styles such as bending the torso backwards. Figure~\ref{fig:Hopper_Vis} shows three policies in a sequence trained using TNB, and all three policies are clearly distinct to one another. More results of different seeds are shown in the video.

\subsubsection{Deceptive Maze}

\begin{figure}[b!]
\vskip -0.1in
\centering
\includegraphics[width=\columnwidth]{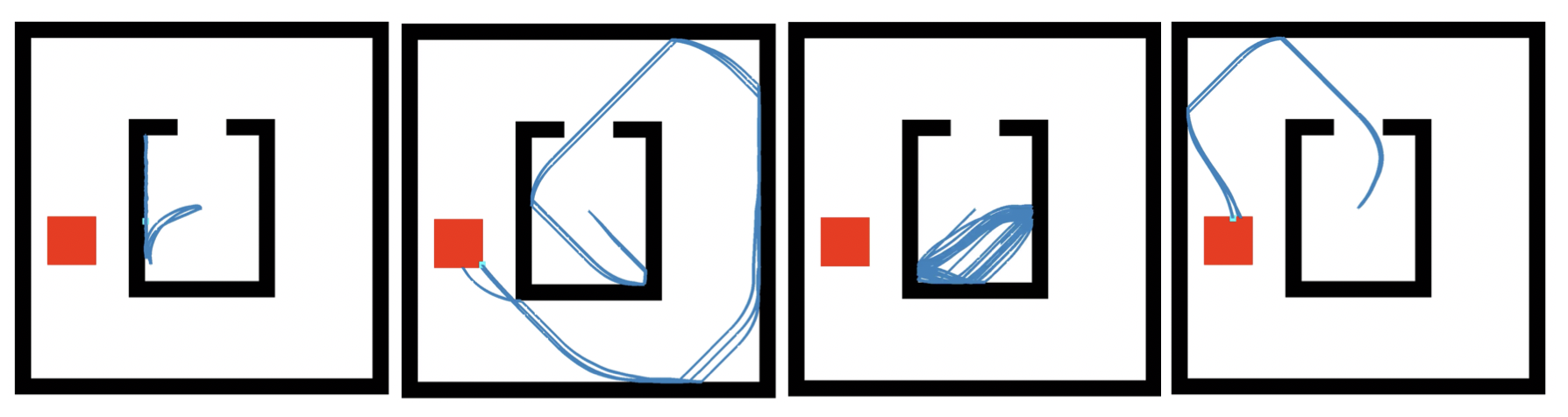}
 \caption{\textbf{Deceptive Maze:} From left to right: four policies sequentially trained using TNB.}
\label{fig:D-Maze_Vis}
\end{figure}

In Deceptive Maze environment (D-Maze), the same observations and actions are used as in the 4-Way Maze problem. In D-Maze, the agent is initially placed inside of an open box, and the goal is placed outside the box. The reward function penalizes the distance between the agent and the goal as well as penalizing the agent for being alive. An episode is terminated early and a large reward is added if the agent reaches the goal. If the agent follows the direction that minimizes distance penalty, the agent will run into a wall and will be blocked from the goal. 

Policies for tasks with deceptive rewards have a clear measure of successes or failures. Thus, we quantitatively measure the performance of this method through two criteria: 1) The average number of successes of a trial, and 2) The number of trials that contain successful policies.

For D-Maze, we run five trials with $k=4$ polices in sequence in each trial for each method. Each policy is trained over 3M samples, and that gives a sample budget of 12M for each trial. As shown in Table~\ref{table:succ}, when running PPO and SAC with different random seeds, very few or none of the policies succeed in driving the agent to the target. In most of the cases, the agent will run straight into the wall on the left shown as the first policy in Figure~\ref{fig:D-Maze_Vis}. When the novelty reward is applied in the training, we see a clear boost in the success rate. The number of policies succeeded are still low when just a weighted sum of two rewards is used (WSR). The poor performance of the weighted sum is possibly due to changes in relative importance of the two rewards for different policies in the sequence, thus making it difficult to find a fixed weight that performs well. With both TNB and TNB-NoProj, on the other hand, both methods are able to give a high success rate and a number of successful policies. 

\begin{table}[h]
\vskip -0.1in
\caption{D-Maze, 12M Sample Budget}
\label{table:succ}
\vskip 0.0in
\begin{center}
\begin{small}
\begin{sc}
\begin{tabular}{lcc}
\toprule
Methods & Avg \# Succ Policies & \# Succ Trials\\
\midrule
SAC     & 0&0/5\\
PPO     & 0.2&1/5\\
WSR    & 0.6&3/5\\
TNB-NoProj    & \textbf{1.2}&3/5\\
TNB    & \textbf{1.2}&\textbf{4/5}\\ 
\bottomrule
\end{tabular}
\end{sc}
\end{small}
\end{center}
\vskip -0.1in
\end{table}

A result of a successful trial using TNB is shown in Figure~\ref{fig:D-Maze_Vis}. We can see that the trails of the policies are significantly different from each other. Although the first policy fails by running straight towards goal, the second policy is encouraged to perform in a different manner and finds a way out the box to reach the goal. Though the the third policy fails, it explores a clearly different set of states from the first two. The fourth policy again finds a solution that is significantly different from the one found in the second policy.

\begin{figure}[b]
\vskip -0.1in
\centering
\includegraphics[width=\columnwidth]{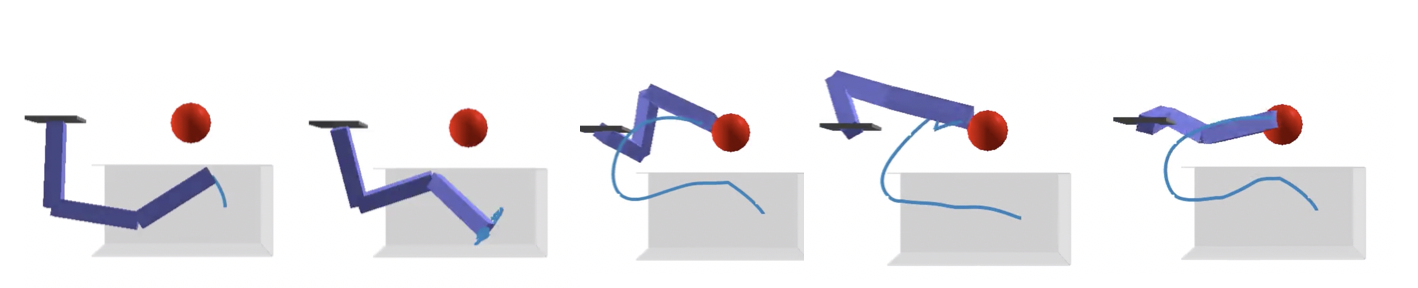}
\caption{\textbf{Deceptive Reacher:} From left to right: five policies sequentially trained using TNB. The first two policies fail, but they guide future policies 3, 4 and 5 to be successful.}
\label{fig:D-Reacher_Vis}
\end{figure}

\subsubsection{Deceptive Reacher}

The Deceptive Reacher (D-Reacher) is a variant of the 3d-Reacher with obstacles and misleading rewards. The end effector of the reacher is initially placed inside a box, and the target position is placed outside the box. The D-Reacher shares the same rewards as the Reacher for the distance to goal and the torque consumption terms. In addition, the D-Reacher has an extra reward component for being alive, and it does not receive a final large reward when finishing the task. By strictly following the reward, the reacher will be trapped by the walls of the box.

We run 5 trials with $k=5$ sequential policies in each trial. Each policy is trained over 6M samples, and that gives 30M sample budget to get five policies in the trial. The performance for D-Reacher is shown in Table~\ref{table:succ}. There is a clear boost of performance after using the novelty reward over plain PPO. There is also a larger number of successful policies in both methods that use gradient bisectors (TNB and TNB-NoProj) in contrast to weighted sums. 

The visual quality of the policies from D-Reacher are shown by plotting the trail of the reacher's end effector. In Figure~\ref{fig:D-Reacher_Vis}, we show five sequential policies from a trial of TNB. The first and second policies fail to move the end effector out of the box, but both explore different states. The remaining three policies all succeed in moving out of the box and solving the task. Moreover, they solve the tasks in three different ways.

\begin{table}[h]
\vskip -0.1in
\caption{D-Reacher, 30M Sample Budget}
\label{table:succ}
\vskip 0.0in
\begin{center}
\begin{small}
\begin{sc}
\begin{tabular}{lcc}
\toprule
Methods & Avg \# Succ Policies & \# Succ Trials\\
\midrule
SAC    & 0&0/5\\
PPO    & 0.2&1/5\\
WSR     & 1.2&\textbf{4/5}\\
TNB-NoProj    & \textbf{2.6}&\textbf{4/5}\\
TNB    & 2.4&\textbf{4/5}\\
\bottomrule
\end{tabular}
\end{sc}
\end{small}
\end{center}
\vskip -0.2in
\end{table}


%% file: Discussion.tex
\section{Discussion}

The above experiments demonstrate using the Task-Novelty Bisector and the Weighted Sum of Rewards approaches to produce novel policies.  Both of these methods proved to be more effective at producing novel policies than simply initializing the policy using different random seeds.  Notably, however, the TNB approach successfully learned novel policies for all five tasks with very few changes for the parameters of the algorithm. The Weighted Sum of the task and novelty rewards required tuning the weights between these two terms in order to produce successful novel policies.  When the weights were not tuned properly, the WSR approach often produced policies that could only achieve one objective over the other.  For this reason, the Task-Novelty Bisector approach is the superior method.

It may be tempting to consider using option discovery approaches such as DIAYN \cite{diayn} or VALOR \cite{valor} to learn novel policies.  There is certainly a similarity between option discovery and finding novel policies, since in both cases the goal is to find a collection of policies that exhibit different sequences of states from one another.  However, current option discovery approaches are not guided by a task reward function.  The policies (options) that they produce do not necessarily satisfy a given task.  For example, only some of the hopper policies produced by DIAYN move forward, and often produce policies that hop in place or even move backwards.  In contrast, the Task-Novelty Bisector method produces hopper policies that always move forward due to the encouragement of forward motion from the task reward function.

It is worth reflecting on why the Task-Novelty Bisector approach is often able to find policies that solve a given task, despite the fact that it does not directly follow the task gradient vector.  The key is that the TNB algorithm modifies the policy parameters using a vector that is still in an uphill direction with respect to the task objective function, just not in the \textit{steepest} uphill direction.  Even when the task and novelty gradients are pointing away from each other, TNB does not move downhill with respect to the task objective.  Nevertheless, it is important to recognize that no policy gradient algorithm can be \textit{guaranteed} to solve a given problem. Even state-of-the-art algorithms such as PPO and SAC can still get caught in local minima and fail to solve a given task. This can be seen in the results of the deceptive reacher and the deceptive maze problems. As described in the Experiments section, often PPO and SAC fail to find a policy that solves these deceptive tasks.

\textbf{Limitations:}  Despite the success of TNB in the environments from the previous section, there are situations where TNB may fail to find suitable novel policies. One such situation is where two successful policies share a common behavior at the start of their rollouts. After the first of these policies is discovered, our novelty reward function will discourage the second policy from being discovered.  An example of this would be a maze in which two different successful paths overlap near the start. It is possible that a larger L might be helpful with this limitation since this would give a larger window when determining the similarity between two segments. However, this could potentially give a longer delay to the novelty reward, making it more difficult for RL algorithms to pick up the reward signal.

Our approach to discovering novel policies is only suitable in the case where there are likely to be a small number of different successful policies.  For a non-deceptive problem where there is just a single successful policy to be found (e.g. a maze with one correct path), our approach may not be helpful. At the other extreme, when there are numerous possible successful policies, it may not be necessary to encourage novelty and using different random seeds may be sufficient.

%% file: Conclusion.tex
\section{Conclusion}

In this work, we have investigated the problem of finding novel policies in a reinforcement learning setting. We demonstrate that the Task-Novelty Bisector approach is an effective method of finding a collection of novel policies for a given task. Novel policies may prove useful in settings where different styles of motion are required, or when a task is deceptive. Because the autoencoder training for TNB is fast, this algorithm provides a compute-efficient alternative to PPO when such variations are desired.

%% file: Acknowledgement.tex
\section{Acknowledgements}
 We thank Karen Liu, Charlie Kemp, Alexander Clegg, Henry Clever and Zackory Erickson for helpful discussions.  This work was supported by NSF award IIS-1514258.

%% file: zAppendix.tex
\appendix
\section{TNB Hyperparameters}
We use $L=45$ and $G=3$ for segmenting the rollouts into training data for autoencoders. We used $G=2$ for the Reacher and 4-Way Maze environments as exceptions because the best policy takes on average less than $45$ steps. When computing the novelty reward using Equation 3 in the main text, we use $w_{novel}=2$ across all examples.

\section{Autoencoder Training}
\label{app:ae_train}
For all of our examples, we use autoencoders that are represented by fully connected neural network with 11 hidden layers consisting of \{1024, 512, 256, 128, 64, 32, 64, 128, 256, 512, 1024\} nodes. We use ReLUs as activation functions for the hidden layers, and linear activation for the output layer. The intuition behind using such a large neural network architecture is that we want the autoencoders to be slightly overfitted to the training data so that it will not accidentally generalize to behaviors that we may deem novel. Each of the autoencoder takes 200 epochs to train with a batch size of 1024 samples, and the training is optimized using the Adam optimizer with learning rate of $10^{-3}$. 

For each policy generated by TNB, we take $10$ policies from the last $50$ iterations of policy training to generate data for training the autoencoders. For each of these policies, we collect $100$ rollouts for data parsing.

\section{PPO Training}
For all experiments in the paper, we use fully connected neural network policies with three 64-unit hidden layers using tanh nonlinearities. During each PPO iteration, we collect $12,000$ steps from the current policy and update the policy parameters using Adam for $3$ epochs, with a mini-batch size of $64$. The learning rate of Adam is set to $0.003$. As suggested in the OpenAI Baselines \cite{dhariwal2017openai} implementation, we use $0.2$ as the clip parameter and 0 for the entropy term in the surrogate loss. For the parameters in the GAE \cite{gae}, we also use the default values in their implementation where $\gamma=0.99$, and $\lambda=0.95$.

\section{SAC Training}
We used the OpenAI Spinning Up implementation for the Soft-Actor-Critic algorithm. For a fair comparison, we used the same number of samples per iterations as in PPO. In particular, we set the replay buffer size to $10^6$, and $\alpha=0.2$ for all tasks.

\section{4-Way Maze Reward Details}
\label{app:4-Way_Maze_rwd}
\begin{figure}[h]
\centering
\includegraphics[height=0.3\textwidth]{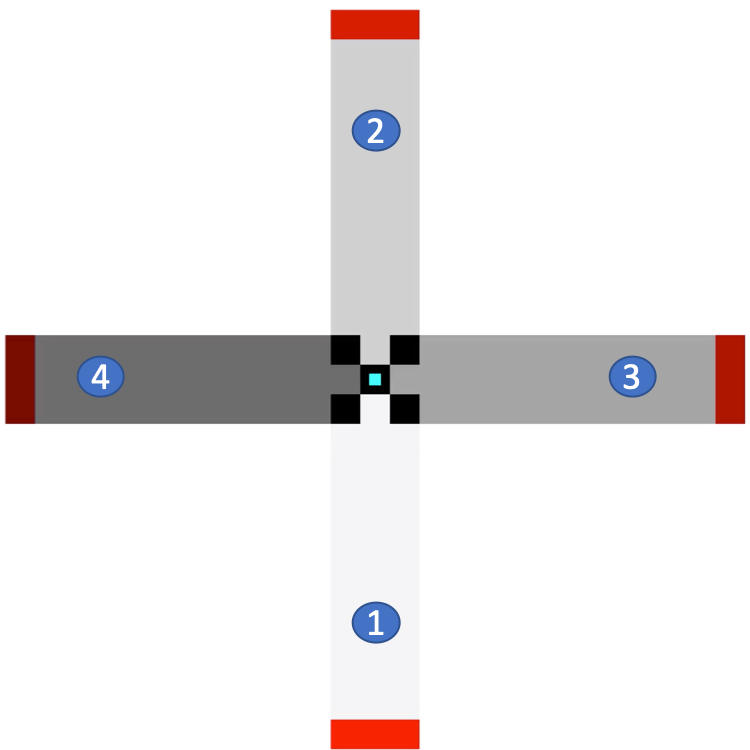}
\caption{Paths in 4-Way Maze are labeled in the decreasing order of rewards.}
\label{fig:4-Way_Maze_Labeled}
\end{figure}

In the 4-Way Maze environment, the agent is given the task to reach one of the four red goals, thus there are four possible solutions to this task. Furthermore, we design a reward function such that different solutions yield different expected rewards. This helps test to see if an algorithm strikes the right balance between finding a novel behavior and obtaining the highest reward. Specifically, the reward signal for reaching a floor cell and a goal cell are formulated as follows:
\begin{equation}
    r^{floor}_i = 50*(\frac{5-i}{4})^3
\end{equation}
\begin{equation}
    r^{goal}_i = 500*(\frac{5-i}{4})^3,
\end{equation}
where $i$ corresponds to the path that has the $i_{th}$ highest expected reward. The cubic order decrease in the reward value aims to create an obvious separation of rewards between paths, so that certain paths are more easily found by an algorithm. In addition to the path rewards, the environment also gives an alive penalty $r^{alive}=-1$ and a penalty of $r^{wall}=-10$ when the agent runs into the wall. An illustration of the 4-way Maze environment is shown in Figure~\ref{fig:4-Way_Maze_Labeled}, where higher reward paths are assigned brighter colors.

\begin{table}[!t]
\caption{D-Maze with different weights}
\label{table:D-Maze_weights}
\vskip 0.15in
\begin{center}
\begin{small}
\begin{sc}
\begin{tabular}{lcc}
\toprule
Weights & Avg \# Succ Policies & \# Succ Trials\\
\midrule
100    & 0.4&1/5\\
200     & 0.6&3/5\\
\textbf{500}    & \textbf{0.6}&\textbf{3/5}\\
1000    & 0.4&2/5\\
\bottomrule
\end{tabular}
\end{sc}
\end{small}
\end{center}
\vskip -0.2in
\end{table}

\begin{table}[!t]
\caption{D-Reacher with different weights}
\label{table:D-Reacher_weights}
\vskip 0.15in
\begin{center}
\begin{small}
\begin{sc}
\begin{tabular}{lcc}
\toprule
Weights & Avg \# Succ Policies & \# Succ Trials\\
\midrule
100    & 1.4&3/5\\
\textbf{200}     & \textbf{1.2}&\textbf{4/5}\\
500    & 1.2&3/5\\
1000    & 1.2&3/5\\
\bottomrule
\end{tabular}
\end{sc}
\end{small}
\end{center}
\vskip -0.2in
\end{table}

\begin{table}[!t]
\vskip 0.15in
\hskip -0.2in
\caption{4-Way Maze with different weights}
\begin{center}
\begin{small}
\begin{sc}
\begin{tabular}{lcc}
\toprule
Weights & Avg \# Paths Explored& \# Trials In Order\\
\midrule
\textbf{100}    & \textbf{3.8}& \textbf{3/5}\\
200    &4 &2/5\\
500    &4 &0/5\\
1000   &4 &0/5\\
\bottomrule
\end{tabular}
\end{sc}
\end{small}
\end{center}
\vskip -0.1in
\label{table:4-Way_Maze}
\end{table}
\section{More Results for Weighted Sum Reward}
\label{app:weights}
For the weighted sum methods, we test multiple weights and report the best performing one. Specifically, we used $100$, $200$, $500$, and $1000$ as our weights to the novelty reward term. We picked this range so that the two rewards shares comparable magnitudes. The bold weights shown in Table~\ref{table:D-Maze_weights}, Table~\ref{table:D-Reacher_weights}, Table~\ref{table:4-Way_Maze} are the weights used for comparison in the paper.

%% file: main.bbl
\begin{thebibliography}{36}
\providecommand{\natexlab}[1]{#1}
\providecommand{\url}[1]{\texttt{#1}}
\expandafter\ifx\csname urlstyle\endcsname\relax
  \providecommand{\doi}[1]{doi: #1}\else
  \providecommand{\doi}{doi: \begingroup \urlstyle{rm}\Url}\fi

\bibitem[Achiam et~al.(2018)Achiam, Edwards, Amodei, and Abbeel]{valor}
Achiam, J., Edwards, H.~A., Amodei, D., and Abbeel, P.
\newblock Variational option discovery algorithms.
\newblock \emph{CoRR}, abs/1807.10299, 2018.

\bibitem[Bacon et~al.(2017)Bacon, Harb, and Precup]{option_critic}
Bacon, P.-L., Harb, J., and Precup, D.
\newblock The option-critic architecture.
\newblock In \emph{AAAI}, pp.\  1726--1734, 2017.

\bibitem[Bellemare et~al.(2016)Bellemare, Srinivasan, Ostrovski, Schaul,
  Saxton, and Munos]{bellemare2016unifying}
Bellemare, M., Srinivasan, S., Ostrovski, G., Schaul, T., Saxton, D., and
  Munos, R.
\newblock Unifying count-based exploration and intrinsic motivation.
\newblock In \emph{Advances in Neural Information Processing Systems}, pp.\
  1471--1479, 2016.

\bibitem[Brockman et~al.(2016)Brockman, Cheung, Pettersson, Schneider,
  Schulman, Tang, and Zaremba]{brockman2016openai}
Brockman, G., Cheung, V., Pettersson, L., Schneider, J., Schulman, J., Tang,
  J., and Zaremba, W.
\newblock Openai gym.
\newblock \emph{arXiv preprint arXiv:1606.01540}, 2016.

\bibitem[Colas et~al.(2018)Colas, Sigaud, and Oudeyer]{colas2018gep}
Colas, C., Sigaud, O., and Oudeyer, P.-Y.
\newblock Gep-pg: Decoupling exploration and exploitation in deep reinforcement
  learning algorithms.
\newblock \emph{arXiv preprint arXiv:1802.05054}, 2018.

\bibitem[Conti et~al.(2018)Conti, Madhavan, Such, Lehman, Stanley, and
  Clune]{conti2018improving}
Conti, E., Madhavan, V., Such, F.~P., Lehman, J., Stanley, K., and Clune, J.
\newblock Improving exploration in evolution strategies for deep reinforcement
  learning via a population of novelty-seeking agents.
\newblock In \emph{Advances in Neural Information Processing Systems}, pp.\
  5027--5038, 2018.

\bibitem[Dhariwal et~al.(2017)Dhariwal, Hesse, Klimov, Nichol, Plappert,
  Radford, Schulman, Sidor, and Wu]{dhariwal2017openai}
Dhariwal, P., Hesse, C., Klimov, O., Nichol, A., Plappert, M., Radford, A.,
  Schulman, J., Sidor, S., and Wu, Y.
\newblock Openai baselines.
\newblock \emph{GitHub, GitHub repository}, 2017.

\bibitem[Eysenbach et~al.(2018)Eysenbach, Gupta, Ibarz, and Levine]{diayn}
Eysenbach, B., Gupta, A., Ibarz, J., and Levine, S.
\newblock Diversity is all you need: Learning diverse skills without a reward
  function.
\newblock 2018.

\bibitem[Gregor et~al.(2016)Gregor, Rezende, and Wierstra]{vic}
Gregor, K., Rezende, D.~J., and Wierstra, D.
\newblock Variational intrinsic control.
\newblock \emph{arXiv preprint arXiv:1611.07507}, 2016.

\bibitem[Haarnoja et~al.(2017)Haarnoja, Tang, Abbeel, and
  Levine]{haarnoja2017reinforcement}
Haarnoja, T., Tang, H., Abbeel, P., and Levine, S.
\newblock Reinforcement learning with deep energy-based policies.
\newblock \emph{arXiv preprint arXiv:1702.08165}, 2017.

\bibitem[Haarnoja et~al.(2018{\natexlab{a}})Haarnoja, Zhou, Abbeel, and
  Levine]{sac}
Haarnoja, T., Zhou, A., Abbeel, P., and Levine, S.
\newblock Soft actor-critic: Off-policy maximum entropy deep reinforcement
  learning with a stochastic actor.
\newblock \emph{arXiv preprint arXiv:1801.01290}, 2018{\natexlab{a}}.

\bibitem[Haarnoja et~al.(2018{\natexlab{b}})Haarnoja, Zhou, Hartikainen,
  Tucker, Ha, Tan, Kumar, Zhu, Gupta, Abbeel, et~al.]{haarnoja2018soft}
Haarnoja, T., Zhou, A., Hartikainen, K., Tucker, G., Ha, S., Tan, J., Kumar,
  V., Zhu, H., Gupta, A., Abbeel, P., et~al.
\newblock Soft actor-critic algorithms and applications.
\newblock \emph{arXiv preprint arXiv:1812.05905}, 2018{\natexlab{b}}.

\bibitem[Hausman et~al.(2018)Hausman, Springenberg, Wang, Heess, and
  Riedmiller]{hausman2018learning}
Hausman, K., Springenberg, J.~T., Wang, Z., Heess, N., and Riedmiller, M.
\newblock Learning an embedding space for transferable robot skills.
\newblock 2018.

\bibitem[Hawkins et~al.(2002)Hawkins, He, Williams, and
  Baxter]{hawkins2002outlier}
Hawkins, S., He, H., Williams, G., and Baxter, R.
\newblock Outlier detection using replicator neural networks.
\newblock In \emph{International Conference on Data Warehousing and Knowledge
  Discovery}, pp.\  170--180. Springer, 2002.

\bibitem[Henderson et~al.(2017)Henderson, Islam, Bachman, Pineau, Precup, and
  Meger]{henderson2017deep}
Henderson, P., Islam, R., Bachman, P., Pineau, J., Precup, D., and Meger, D.
\newblock Deep reinforcement learning that matters.
\newblock \emph{arXiv preprint arXiv:1709.06560}, 2017.

\bibitem[Hester \& Stone(2017)Hester and Stone]{hester2017intrinsically}
Hester, T. and Stone, P.
\newblock Intrinsically motivated model learning for developing curious robots.
\newblock \emph{Artificial Intelligence}, 247:\penalty0 170--186, 2017.

\bibitem[Houthooft et~al.(2016)Houthooft, Chen, Duan, Schulman, De~Turck, and
  Abbeel]{Vime}
Houthooft, R., Chen, X., Duan, Y., Schulman, J., De~Turck, F., and Abbeel, P.
\newblock Vime: Variational information maximizing exploration.
\newblock In \emph{Advances in Neural Information Processing Systems}, pp.\
  1109--1117, 2016.

\bibitem[Lee et~al.(2018)Lee, Grey, Ha, Kunz, Jain, Ye, Srinivasa, Stilman, and
  Liu]{lee2018dart}
Lee, J., Grey, M.~X., Ha, S., Kunz, T., Jain, S., Ye, Y., Srinivasa, S.~S.,
  Stilman, M., and Liu, C.~K.
\newblock Dart: Dynamic animation and robotics toolkit.
\newblock \emph{The Journal of Open Source Software}, 3\penalty0 (22):\penalty0
  500, 2018.

\bibitem[Lehman \& Stanley(2011{\natexlab{a}})Lehman and Stanley]{evoNovel}
Lehman, J. and Stanley, K.~O.
\newblock Abandoning objectives: Evolution through the search for novelty
  alone.
\newblock \emph{Evolutionary computation}, 19\penalty0 (2):\penalty0 189--223,
  2011{\natexlab{a}}.

\bibitem[Lehman \& Stanley(2011{\natexlab{b}})Lehman and
  Stanley]{lehman2011evolving}
Lehman, J. and Stanley, K.~O.
\newblock Evolving a diversity of virtual creatures through novelty search and
  local competition.
\newblock In \emph{Proceedings of the 13th annual conference on Genetic and
  evolutionary computation}, pp.\  211--218. ACM, 2011{\natexlab{b}}.

\bibitem[Liao \& Vemuri(2002)Liao and Vemuri]{liao2002use}
Liao, Y. and Vemuri, V.~R.
\newblock Use of k-nearest neighbor classifier for intrusion detection.
\newblock \emph{Computers \& security}, 21\penalty0 (5):\penalty0 439--448,
  2002.

\bibitem[Lillicrap et~al.(2015)Lillicrap, Hunt, Pritzel, Heess, Erez, Tassa,
  Silver, and Wierstra]{ddpg}
Lillicrap, T.~P., Hunt, J.~J., Pritzel, A., Heess, N., Erez, T., Tassa, Y.,
  Silver, D., and Wierstra, D.
\newblock Continuous control with deep reinforcement learning.
\newblock \emph{arXiv preprint arXiv:1509.02971}, 2015.

\bibitem[Mouret \& Doncieux(2009)Mouret and Doncieux]{mouret2009overcoming}
Mouret, J.-B. and Doncieux, S.
\newblock Overcoming the bootstrap problem in evolutionary robotics using
  behavioral diversity.
\newblock In \emph{Evolutionary Computation, 2009. CEC'09. IEEE Congress on},
  pp.\  1161--1168. IEEE, 2009.

\bibitem[Pugh et~al.(2016)Pugh, Soros, and Stanley]{pugh2016quality}
Pugh, J.~K., Soros, L.~B., and Stanley, K.~O.
\newblock Quality diversity: A new frontier for evolutionary computation.
\newblock \emph{Frontiers in Robotics and AI}, 3:\penalty0 40, 2016.

\bibitem[Richter \& Roy(2017)Richter and Roy]{richter2017safe}
Richter, C. and Roy, N.
\newblock Safe visual navigation via deep learning and novelty detection.
\newblock 2017.

\bibitem[Ruff et~al.(2018)Ruff, G{\"o}rnitz, Deecke, Siddiqui, Vandermeulen,
  Binder, M{\"u}ller, and Kloft]{ruff2018deep}
Ruff, L., G{\"o}rnitz, N., Deecke, L., Siddiqui, S.~A., Vandermeulen, R.,
  Binder, A., M{\"u}ller, E., and Kloft, M.
\newblock Deep one-class classification.
\newblock In \emph{International Conference on Machine Learning}, pp.\
  4390--4399, 2018.

\bibitem[Schmidhuber(1991)]{schmidhuber1991curious}
Schmidhuber, J.
\newblock Curious model-building control systems.
\newblock In \emph{Neural Networks, 1991. 1991 IEEE International Joint
  Conference on}, pp.\  1458--1463. IEEE, 1991.

\bibitem[Schulman et~al.(2015{\natexlab{a}})Schulman, Levine, Abbeel, Jordan,
  and Moritz]{trpo}
Schulman, J., Levine, S., Abbeel, P., Jordan, M., and Moritz, P.
\newblock Trust region policy optimization.
\newblock In \emph{International Conference on Machine Learning}, pp.\
  1889--1897, 2015{\natexlab{a}}.

\bibitem[Schulman et~al.(2015{\natexlab{b}})Schulman, Moritz, Levine, Jordan,
  and Abbeel]{gae}
Schulman, J., Moritz, P., Levine, S., Jordan, M., and Abbeel, P.
\newblock High-dimensional continuous control using generalized advantage
  estimation.
\newblock \emph{arXiv preprint arXiv:1506.02438}, 2015{\natexlab{b}}.

\bibitem[Schulman et~al.(2017)Schulman, Wolski, Dhariwal, Radford, and
  Klimov]{ppo}
Schulman, J., Wolski, F., Dhariwal, P., Radford, A., and Klimov, O.
\newblock Proximal policy optimization algorithms.
\newblock \emph{arXiv preprint arXiv:1707.06347}, 2017.

\bibitem[Strehl \& Littman(2005)Strehl and Littman]{strehl2005theoretical}
Strehl, A.~L. and Littman, M.~L.
\newblock A theoretical analysis of model-based interval estimation.
\newblock In \emph{Proceedings of the 22nd international conference on Machine
  learning}, pp.\  856--863. ACM, 2005.

\bibitem[Sun et~al.(2011)Sun, Gomez, and Schmidhuber]{sun2011planning}
Sun, Y., Gomez, F., and Schmidhuber, J.
\newblock Planning to be surprised: Optimal bayesian exploration in dynamic
  environments.
\newblock In \emph{International Conference on Artificial General
  Intelligence}, pp.\  41--51. Springer, 2011.

\bibitem[Sutton \& Barto(2018)Sutton and Barto]{sutton2018reinforcement}
Sutton, R.~S. and Barto, A.~G.
\newblock \emph{Reinforcement learning: An introduction}.
\newblock MIT press, 2018.

\bibitem[Sutton et~al.(1999)Sutton, Precup, and Singh]{optionFramework}
Sutton, R.~S., Precup, D., and Singh, S.
\newblock Between mdps and semi-mdps: A framework for temporal abstraction in
  reinforcement learning.
\newblock \emph{Artificial intelligence}, 112\penalty0 (1-2):\penalty0
  181--211, 1999.

\bibitem[Tang et~al.(2017)Tang, Houthooft, Foote, Stooke, Chen, Duan, Schulman,
  DeTurck, and Abbeel]{tang2017exploration}
Tang, H., Houthooft, R., Foote, D., Stooke, A., Chen, O.~X., Duan, Y.,
  Schulman, J., DeTurck, F., and Abbeel, P.
\newblock \# exploration: A study of count-based exploration for deep
  reinforcement learning.
\newblock In \emph{Advances in neural information processing systems}, pp.\
  2753--2762, 2017.

\bibitem[Zhao \& Saligrama(2009)Zhao and Saligrama]{zhao2009anomaly}
Zhao, M. and Saligrama, V.
\newblock Anomaly detection with score functions based on nearest neighbor
  graphs.
\newblock In \emph{Advances in neural information processing systems}, pp.\
  2250--2258, 2009.

\end{thebibliography}
